\documentclass[conference,a4paper]{IEEEtran}
\IEEEoverridecommandlockouts

\usepackage[T1]{fontenc}    
\usepackage[utf8]{inputenc} 

\usepackage[hidelinks]{hyperref}
\usepackage[cmex10]{amsmath}
\usepackage{amssymb,amsfonts}
\usepackage{dblfloatfix}
\usepackage{enumitem}

\usepackage{mwe}

\usepackage[ruled,vlined]{algorithm2e}
\usepackage{graphicx}
\graphicspath{{fig/}}

\usepackage{booktabs}
\usepackage{siunitx}
\usepackage[numbers,compress]{natbib}
\usepackage{texnames}
\usepackage{bm,bbm}
\usepackage{orcidlink}

\usepackage{xcolor}

\usepackage{caption}

\newcommand{\para}[1]{\noindent\textbf{#1.}\ }

\begin{document}

\title{\uppercase{Flood Mapping from RGB imagery using a Vision Foundation Model}\vspace*{-0.3 cm}

}

\author{
\IEEEauthorblockN{
Vladyslav Polushko\IEEEauthorrefmark{1}\IEEEauthorrefmark{2},
Tilman Bucher\IEEEauthorrefmark{3},
Ronald R\"osch\IEEEauthorrefmark{1},
Thomas M\"arz\IEEEauthorrefmark{2},
Markus Rauhut\IEEEauthorrefmark{1},
Andreas Weinmann\IEEEauthorrefmark{4}
}\\[-0.9em]
\IEEEauthorblockA{
        \IEEEauthorrefmark{1}Image Processing Department, 
        Fraunhofer ITWM,
        Kaiserslautern, Germany,\\
        \{vladyslav.polushko, ronald.roesch, markus.rauhut\}@itwm.fraunhofer.de
        }\\[-0.9em]
\IEEEauthorblockA{
        \IEEEauthorrefmark{2}ACIDA Lab, 
        Hochschule Darmstadt,
        Darmstadt, Germany.
        \{vladyslav.polushko, thomas.maerz\}@h-da.de
        }\\[-0.9em]
\IEEEauthorblockA{
        \IEEEauthorrefmark{3} 
        Institut f{\"u}r Optische Sensorsysteme, DLR,
        Berlin, Germany,
        tilman.bucher@dlr.de
        }\\[-0.9em]
\IEEEauthorblockA{%
  \IEEEauthorrefmark{4}
  		ACIDA Lab, 
  		THWS W{\"u}rzburg-Schweinfurt, Schweinfurt, Germany, 
  		andreas.weinmann@thws.de
}
}

\maketitle

\begin{abstract}
Timely, high-resolution maps of flood extent around settlements are essential for emergency response and damage assessment. We consider airborne RGB imagery
as data for flood mapping as it can be collected rapidly 
at low cost.
To produce flood maps, deep learning models for water segmentation are frequently used.
Often, CNN based and small vision transformer models are employed.
However, they need a lot of data for adaptation to a change of scenery, i.e., another flooding event.
Vision foundation models or large vision transformers
are known for their capability to generalize across domains. More recently vision foundation models
for Earth observation became available. They are pretrained on satellite data, whose spatial resolution, viewing geometry, and radiometry differ from nadir RGB imagery. 
Therefore, adaptation is required.
In this study, we investigate how a satellite-pretrained Earth observation foundation 
model can be adapted to centimeter-scale flood-water mapping from RGB imagery. 
Specifically, we fine-tune a model we call Prithvi-2.0-UPN   
which consists of the Prithvi-EO-2.0-600M Vision Transformer combined with a UPerNet 
decoder for binary water segmentation on two different RGB datasets (BlessemFlood21, NeuenahrFlood). 
In a first baseline experiment we observe that Prithvi-2.0-UPN 
reaches state-of-the-art results on BlessemFlood21 and NeuenahrFlood, 
when trained on the respective datasets. 
In a second experiment we show 
that Prithvi-2.0-UPN performs better than state-of-the-art baseline models for transfer application to another flood event
(trained on BlessemFlood21, tested on NeuenahrFlood) in a zero-shot fashion.
However, the performance indicates room for improvement.
In this respect,
we investigate in a third experiment how the performance improves when further fine-tuning 
the models with small shares of NeuenahrFlood training data:
Prithvi-2.0-UPN improves the fastest and reaches almost 
the level of performance when fully trained on NeuenahrFlood, 
indicating transfer capabilities.
\end{abstract}

\begin{IEEEkeywords}
	Remote Sensing, Deep Learning, Vision Foundation Models, Semantic Segmentation, Water Detection
\end{IEEEkeywords}

\section{Introduction}

Flood events occur across the globe and cause substantial danger and damages~\cite{disastercred24,swiss21,ourworld2021online,IPPC21}.
For humanitarian organizations, detailed 
information on flood extent 
is important for effective response.
Beyond satellite-based monitoring, airborne platforms equipped with RGB cameras are increasingly
used to obtain targeted imagery over flood-affected areas~\cite{papyan24ai,daud22applications,ruwanpathirana24evaluation}. Drones can be deployed on short notice, provide imagery at centimeter resolution, and can be georeferenced on-board
or on-site~\cite{reiss2022evaluation,halbgewachs2023using}, so that the data can be for instance aligned with semantic road network
information~\cite{haklay08openstreetmap} which enables 
accessibility analysis around roads and buildings. 
Deep Learning (DL) has become a standard tool for automated analysis of Remote Sensing (RS) data. Numerous works address
flood detection and change mapping~\cite{fernandes22river,Zhao24sarfloods,li24domainfoundationflood,li24overcominguncertaintyflooding}
as well as infrastructure-related tasks such as road and building extraction~\cite{buslaev18fcnroadextraction,hetang24segment,ma2024stripunetroadextraction}.
Flood mapping is typically formulated as a semantic segmentation problem 
for which convolutional and transformer-based architectures have been used.
Examples are DeepLabV3+~\cite{chen18deeplabv3+}, UNet++~\cite{zhou18unet++},
and SegFormer-b5~\cite{xie21segformer}.
Several approaches combine water segmentation with infrastructure semantics, e.g., by
fusing flood masks with road or building data or assigning flood attributes to roads/buildings from pre- and post-event
imagery~\cite{munawar21uavs,nair24enhanced,hansch22spacenet}. High-resolution flood datasets such as FloodNet~\cite{floodnet21}
have been used to benchmark segmentation models for post-disaster assessment~\cite{dixit23post}, building vulnerability~\cite{xing23flood},
and road damage detection and tracking~\cite{zhao22road}. Most of this work focuses on satellite sensors and does not target
centimeter-scale RGB imagery.
In previous work, we introduced two high-resolution airborne labeled datasets for flood-water segmentation,
\textit{BlessemFlood21}~\cite{polushko2024blessemflood21ieee} and \textit{NeuenahrFlood}~\cite{polushko25Neuenahrlood}.
Both contain georeferenced airborne RGB imagery of the floods in Blessem and Bad Neuenahr at \SI{0.15}{\meter}/px. The labels are pixel-wise water masks obtained in human-in-the-loop annotation workflows. On
BlessemFlood21, the baseline models DeepLabV3+~\cite{chen18deeplabv3+}, UNet++~\cite{zhou18unet++},
and SegFormer-b5~\cite{xie21segformer} were trained for flood-water segmentation~\cite{polushko2024blessemflood21ieee}. 
We used the predicted water masks together with OSM road data to identify flooded road segments in~\cite{polushko25floodedroads}. These studies showed that event-specific models can produce
accurate flood-water and flooded-road maps from RGB imagery. 

Earth Observation Foundation Models (EOFMs) have recently been proposed to reduce repeated training effort. Models such as
Prithvi-EO and SatMAE are pretrained on multi-temporal, multi-spectral satellite imagery using self-supervised
objectives and then adapted to downstream tasks via fine-tuning~\cite{jakubik23foundation,szwarcman24prithvi,cong22satmae}.
Benchmark studies~\cite{lacoste23geo,dionelis24evaluating,zhang24geoscience} indicate that EOFMs can provide strong
performance across diverse RS tasks while reducing requirements for labeled data and computational cost. For flood mapping,
recent works have adapted such models to satellite imagery, e.g., by assessing Prithvi-EO for flood inundation mapping from
Sentinel-2 data, studying region-specific adaptation, or proposing hybrid encoder-decoder architectures and parameter-efficient
fine-tuning~\cite{li23assessment,tamura25earthFM,kostejn25uprithvi,selvam24rapidadaption}. These studies show that
satellite-pretrained EOFMs can be used successfully for flood segmentation on optical satellite imagery. However, they remain
confined to satellite benchmarks.
So far, existing EOFM-based work does not consider flood-water segmentation from single-date, very high-resolution non-satellite
RGB imagery such as airborne or drone acquisitions, where labeled datasets are typically small and tied to individual
events and the domain differs markedly from multi-spectral satellite imagery. As a result, so far there is no systematic
assessment of how a satellite-pretrained model such as Prithvi-EO-2.0~\cite{szwarcman24prithvi} performs when adapted to
RGB flood imagery, how much labeled data is required to obtain competitive performance, how robustly such a model
transfers between different flood events, and how quickly useful performance is reached during fine-tuning. Clarifying these
aspects is essential for assessing whether Vision Foundation Models can be used effectively for data-efficient,
high-resolution flood-water segmentation from non-satellite imagery w.r.t. rapid humanitarian response.

\para{Contributions} We address these questions by adapting Prithvi-EO-2.0 to water segmentation using high-resolution 
airborne RGB datasets, and we investigate model transferability between flood events. 
More precisely,
the contributions of this work are:
\begin{enumerate}[label=\roman*)] 
\item We adapt and fine-tune the satellite-pretrained Vision Foundation Model Prithvi-EO-2.0 
with a UPerNet decoder (which we call Prithvi-2.0-UPN) to the high-resolution RGB airborne imagery 
on the BlessemFlood21 and NeuenahrFlood datasets and assess its in-domain 
(training and evaluation on the same flood event) performance relative to baseline models.
\item We investigate the generalization across flood events by two experiments: first, we study 
transfer from BlessemFlood21 to NeuenahrFlood in a zero-shot manner, meaning that the model has 
not seen NeuenahrFlood data before. Secondly, we examine how further fine-tuning with increasing 
small shares of NeuenahrFlood data improves model performance compared with the baseline models. 
This imitates the situation where only a smaller amount of labeled data can be acquired. 
\end{enumerate}

\section{Data}

The task is binary pixel-wise segmentation of flood water versus non-water from RGB imagery. 
For training and testing we use two different high-resolution RGB datasets for flood-water segmentation:
\textit{BlessemFlood21}~\cite{polushko2024blessemflood21ieee} (acquired via gyrocopter)
and
\textit{NeuenahrFlood}~\cite{polushko25Neuenahrlood} (acquired via airplane).
Both datasets consist of contiguous RGB orthomosaics at a ground sampling distance of \SI{0.15}{\meter}/px.
The training and test sets consist of non-overlapping $512\times512$ RGB tiles extracted from the orthomosaics.
Each tile is paired with a binary reference mask indicating flood water
(foreground) and non-water (background).
In both datasets, flooded areas typically occupy only a small portion of each tile.
Tiles with at least \SI{5}{\percent} water pixels in the water mask are
marked as \emph{water images}, the remaining tiles as \emph{non-water images}.
The \SI{5}{\percent} threshold ensures that water images
contain a clearly visible flooded area and are not dominated by background with only a 
few isolated water pixels. 
\textit{BlessemFlood21} 
comprises 4\,623 tiles in total and contains 336 water images, 
while \textit{NeuenahrFlood} comprises 21\,584 tiles in total and 1272 water images. 
To obtain training and test subsets that are stable and representative 
with respect to this imbalance, we construct
train/test splits by water-coverage-based sampling. 
The global proportion of water to non-water images is approximately
preserved so that all subsets reflect the overall scarcity of flooded scenes.
The split is \SI{80}{\percent} training images and \SI{20}{\percent} test images.
For fine-tuning on small shares (see Sect.~\ref{sect:results} Exp.~3),
subsets of 128, 256, 512, 1\,024, and 2\,048 training tiles are drawn from the NeuenahrFlood training data
such that again the proportion is preserved.
%


\section{Model}

We use the Prithvi-EO-2.0-600M Vision Foundation Model with UPerNet~\cite{szwarcman24prithvi,wang23upernet} for flood-water segmentation (which we call Prithvi-2.0-UPN). The encoder is a transformer pretrained in a self-supervised manner on 4.2\,million training samples from Landsat and Sentinel-2 (HLS) optical satellite imagery. The inputs are $512\times512$ RGB tiles, and the model outputs pixel-wise flood probabilities, which are thresholded at 0.5 to obtain binary water masks. As decoder we use the UPerNet framework, see \cite{wang23upernet}.

Encoder plus decoder are fine-tuned end-to-end. The model (as well as its baseline comparison models) is trained in PyTorch with AdamW and a cosine-annealed learning rate schedule over 500 epochs, using a batch size of 12 and a binary focal loss on the flood/non-flood logits. The initial learning rate is $10^{-4}$ for all experiments except Experiment~3, where it is set to $2\cdot10^{-5}$. Unspecified Prithvi hyperparameters were kept at their default values.


\section{Experiments and Results}\label{sect:results}

\paragraph{Experimental design}
We evaluate Prithvi-2.0-UPN and, as baselines, DeepLabV3+, UNet++, and SegFormer-b5 
(cf. \cite{polushko24blessem}) on BlessemFlood21 and NeuenahrFlood in three experimental setups. 
Experiment~1 trains all models on BlessemFlood21 and, separately, on NeuenahrFlood 
and reports the metrics on the respective test sets. 
The metrics we use throughout the paper are Intersection-over-Union (IoU), Dice score, and pixel-wise Accuracy (Acc.).
Scores of the metrics are stated in percent.
Experiment~2 trains the models on BlessemFlood21 and evaluates them without adaptation on 
the NeuenahrFlood test set; in addition to the metrics at a fixed probability threshold, 
we inspect representative probability maps and vary the decision threshold to analyze its 
effect on segmentation performance. 
Experiment~3 initializes Prithvi-2.0-UPN with the BlessemFlood21-trained weights and fine-tunes 
it end-to-end on increasingly large subsets of the NeuenahrFlood training data 
(128, 256, 512, 1\,024, and 2\,048 tiles); for each subset size we evaluate the metrics
on the NeuenahrFlood test set to study how performance scales with an increasing size. 
The same setup is used for  the baseline models. 
Throughout this section, the best values in each table are shown in bold and the second-best in italics.

\paragraph{Experiment 1: Baseline comparison}
We compare all models when training and testing on the same flood dataset, 
using \textit{BlessemFlood21} and \textit{NeuenahrFlood}. DeepLabV3+, UNet++, and SegFormer-b5 
follow the configurations of~\cite{polushko24blessem}. Prithvi-2.0-UPN is fine-tuned 
separately on the BlessemFlood21 and NeuenahrFlood training tiles and evaluated on the corresponding test sets. The goal is to assess whether the satellite-pretrained Prithvi-2.0-UPN reaches the performance level of the baselines.

\begin{figure}[t]
    \centering
    \begin{minipage}[t]{0.166\linewidth}
        \centering
        \includegraphics[width=\linewidth]{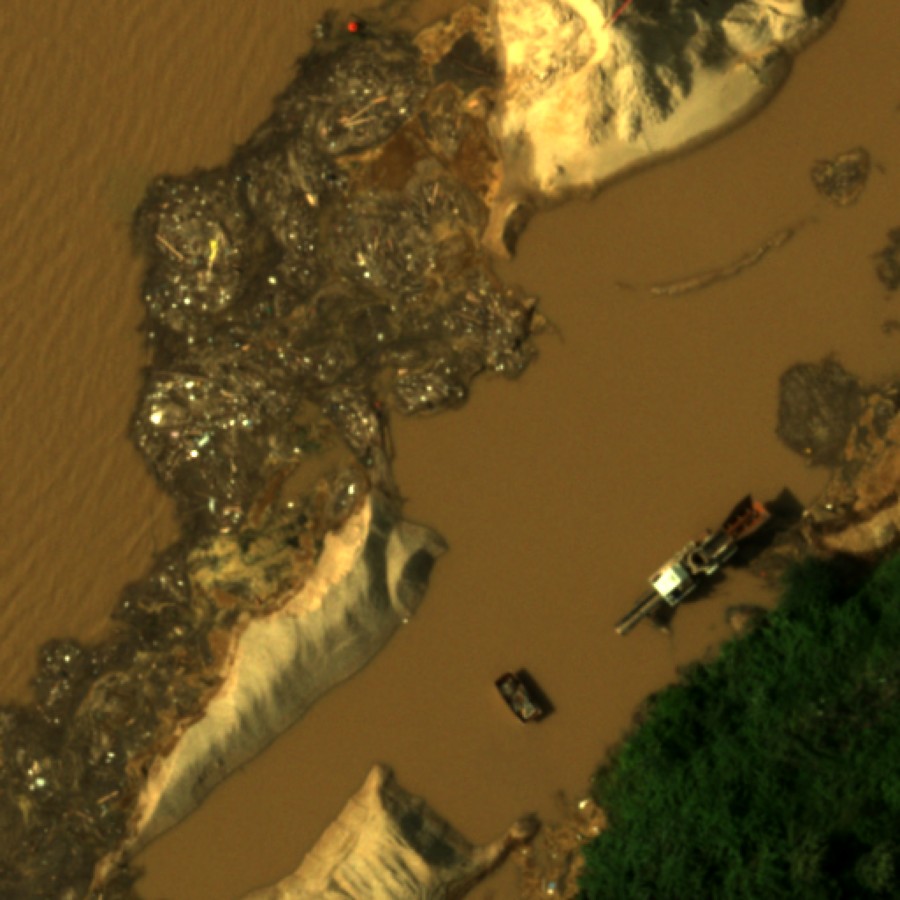}
        \vspace{0.15em}
        \footnotesize (a) RGB
    \end{minipage}\hspace{0.05mm}%
    \begin{minipage}[t]{0.166\linewidth}
        \centering
        \includegraphics[width=\linewidth]{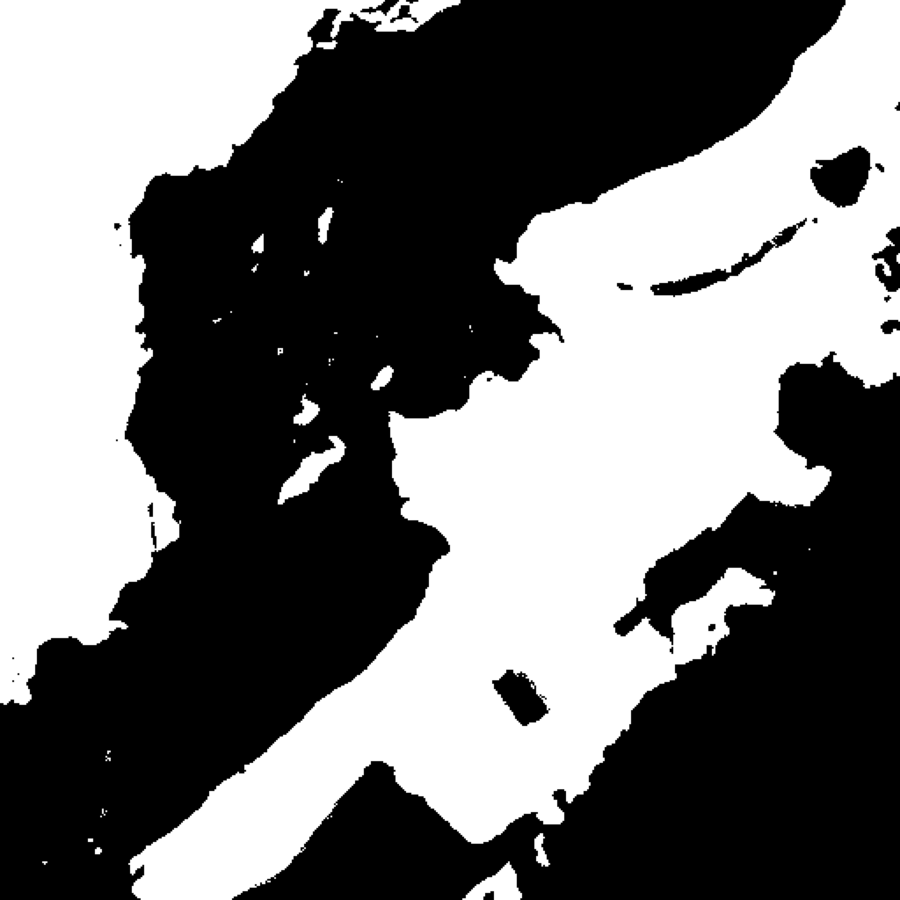}
        \vspace{0.15em}
        \footnotesize (b) GT
    \end{minipage}\hspace{0.05mm}%
    \begin{minipage}[t]{0.166\linewidth}
        \centering
        \includegraphics[width=\linewidth]{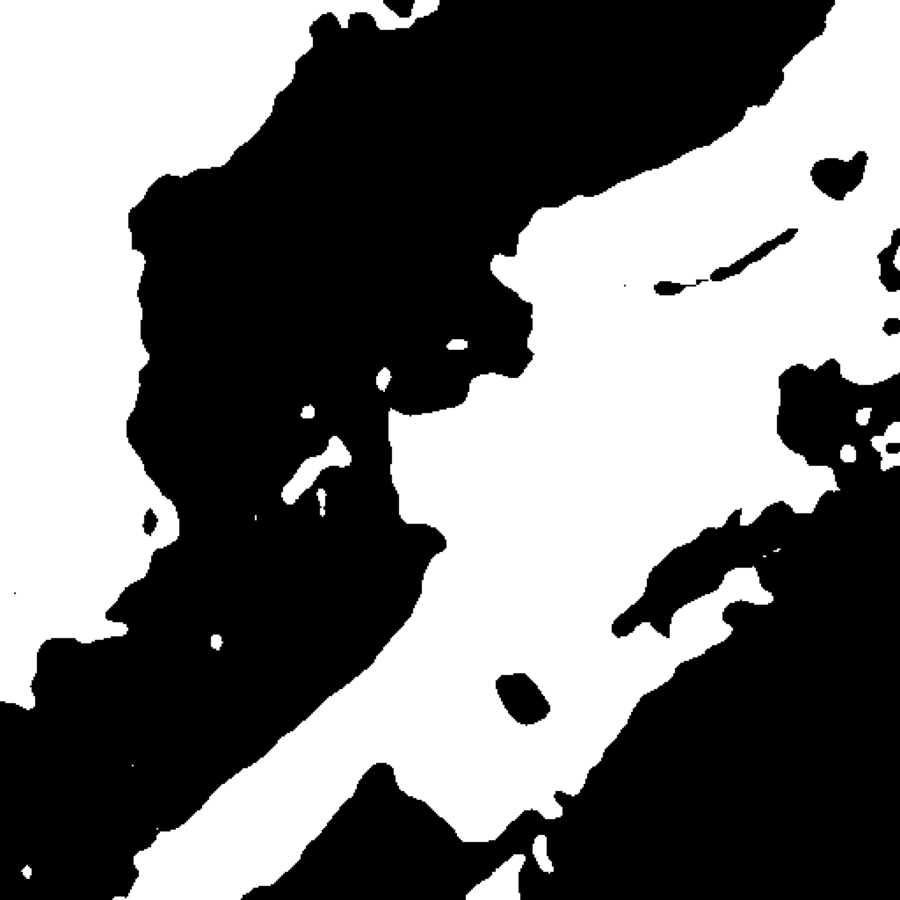}
        \vspace{0.15em}
        \footnotesize (c) Prithvi
    \end{minipage}\hspace{0.05mm}%
    \begin{minipage}[t]{0.166\linewidth}
        \centering
        \includegraphics[width=\linewidth]{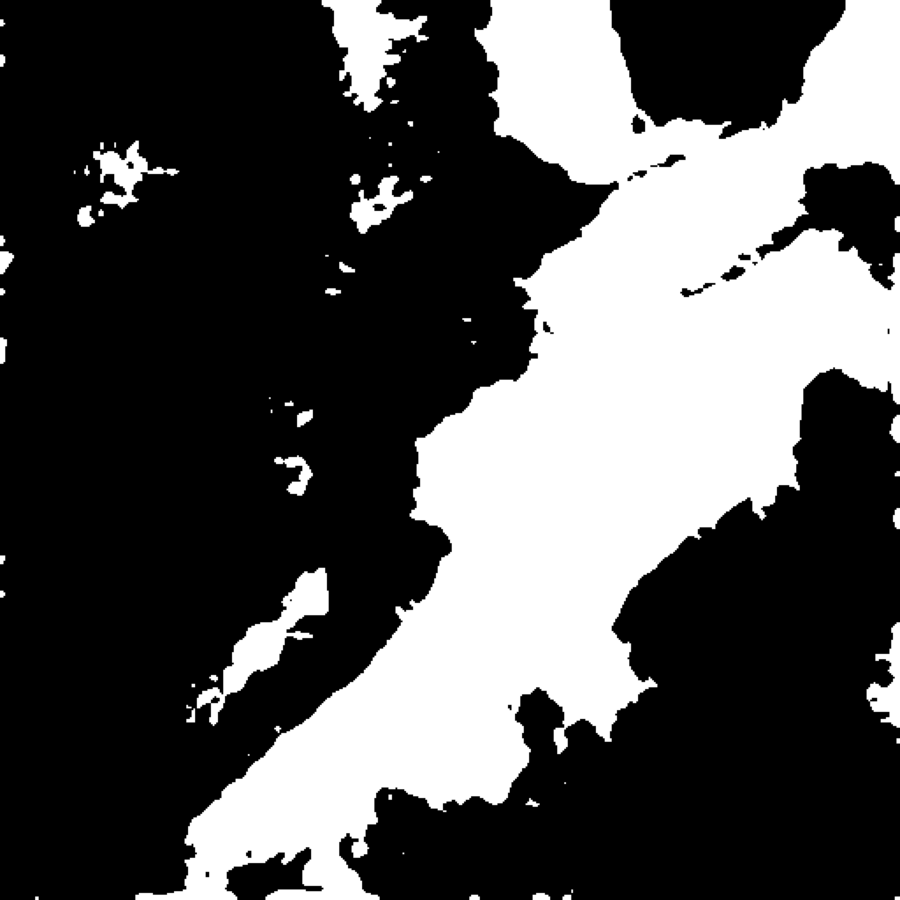}
        \vspace{0.15em}
        \footnotesize (d) SegF
    \end{minipage}\hspace{0.05mm}%
    \begin{minipage}[t]{0.166\linewidth}
        \centering
        \includegraphics[width=\linewidth]{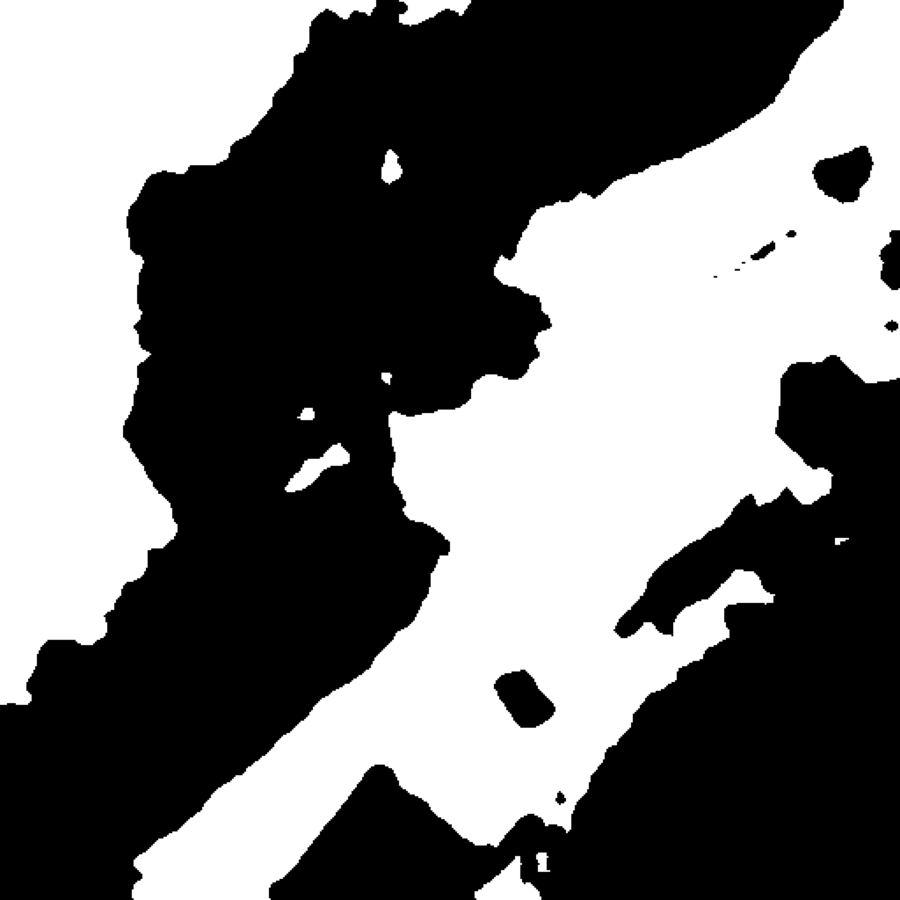}
        \vspace{0.15em}
        \footnotesize (e) DeepLab
    \end{minipage}\hspace{0.05mm}%
    \begin{minipage}[t]{0.166\linewidth}
        \centering
        \includegraphics[width=\linewidth]{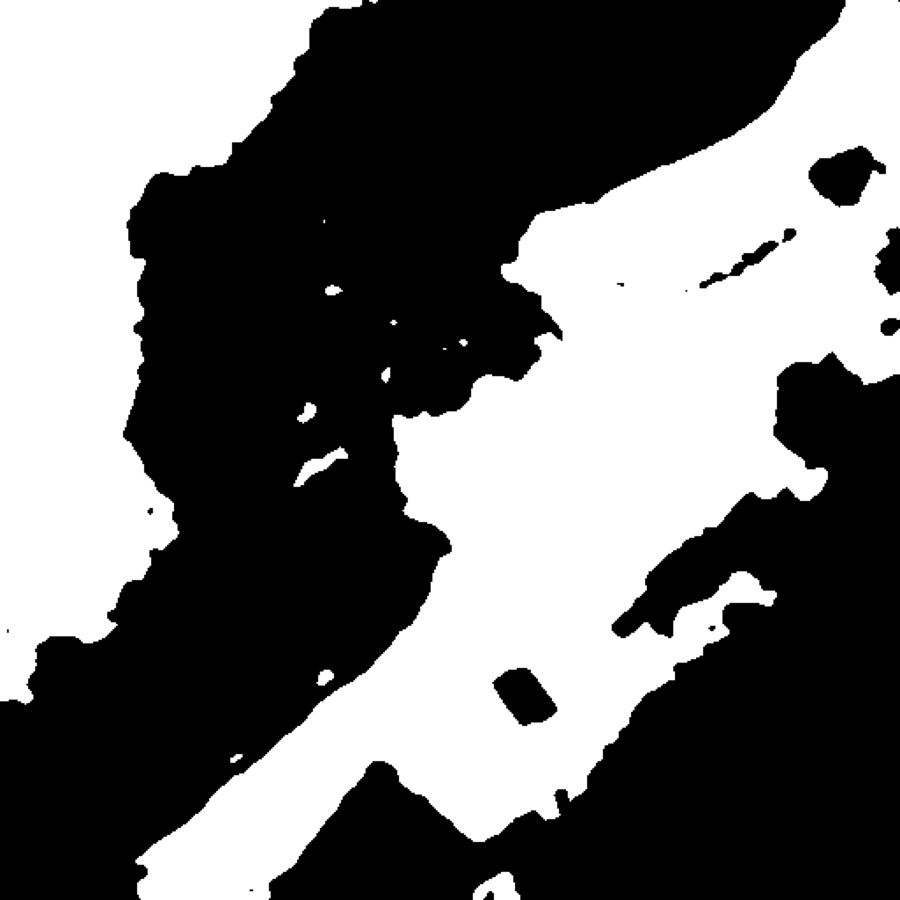}
        \vspace{0.15em}
        \footnotesize (f) UNet++
    \end{minipage}

\caption{Qualitative comparison on a \textit{BlessemFlood21} test tile (RGB input) and predictions from Prithvi-2.0-UPN, and baseline methods; threshold 0.5 (Exp.~1).}
\label{fig:in-domain-performancne}
\end{figure}

Fig.~\ref{fig:in-domain-performancne} shows a qualitative comparison on a \textit{BlessemFlood21} test tile. 
Across all models, the main flooded corridor is recovered as a coherent water body. 
Prithvi-2.0-UPN and DeepLabV3+ follow the shoreline closely and produce smooth flood boundaries, 
whereas SegFormer-b5 and UNet++ show more irregular outlines and small gaps near the water--land boundary. 
Concerning quantitative evaluation on \textit{BlessemFlood21} 
(cf. Tab.~\ref{tab:model_comparison_blessem}), DeepLabV3+ achieves the highest IoU and Dice, 
with Prithvi-2.0-UPN very close behind and with the highest pixel accuracy. 
UNet++ and SegFormer-b5 obtain clearly lower IoU and Dice. 

\begin{table}[ht]
\centering
\small
\setlength{\tabcolsep}{3pt}
\begin{tabular*}{\columnwidth}{@{\extracolsep{\fill}} l l r r r r @{}}
\toprule
Model           & Encoder   & Params & IoU              & Dice             & Acc.        \\
\midrule
DeepLabV3+      & ResNet-50 &  26.7M & \textbf{90.6}    & \textbf{95.1}    & 99.3            \\ 
UNet++          & ResNet-50 &  49.0M & 84.2             & 91.4             & \emph{99.5}     \\ 
SegFormer-b5    & Hier. ViT &  84.6M & 75.4             & 78.4             & 95.5            \\ 
Prithvi-2.0-UPN & ViT       & 600M   & \emph{89.1}      & \emph{94.3}      & \textbf{99.6}   \\
\bottomrule
\end{tabular*}
\caption{Metrics on the \textit{BlessemFlood21} test set (Exp.~1).}
\label{tab:model_comparison_blessem}
\end{table}
On \textit{NeuenahrFlood} 
(cf. Tab.~\ref{tab:model_comparison_ahr}) UNet++ reaches the highest IoU and Dice score. 
SegFormer-b5 and Prithvi-2.0-UPN follow closely, with Prithvi-2.0-UPN achieving the 
highest accuracy. DeepLabV3+ has slightly lower IoU and Dice score. 

\begin{table}[ht]
\centering
\small
\setlength{\tabcolsep}{3pt}
\begin{tabular*}{\columnwidth}{@{\extracolsep{\fill}} l l r r r r @{}}
\toprule
Model           & Encoder   & Params & IoU           & Dice           & Acc.          \\
\midrule
DeepLabV3+      & ResNet-50 &  26.7M & 89.5          & 94.4           & 97.9          \\ 
UNet++          & ResNet-50 &  49.0M & \textbf{93.1} & \textbf{96.4}  & \emph{99.1}   \\ 
SegFormer-b5    & Hier. ViT &  84.6M & 90.2          & \emph{94.8}    & 98.5          \\ 
Prithvi-2.0-UPN & ViT       & 600M   & \emph{90.3}   & 94.6           & \textbf{99.5} \\
\bottomrule
\end{tabular*}
\caption{
Metrics on the  \textit{NeuenahrFlood} test set (Exp.~1).}
\label{tab:model_comparison_ahr}
\end{table}
Overall, Prithvi-2.0-UPN consistently ranks among the top models on both datasets:
with regard to IoU and Dice it keeps up with  
the best baseline models (which are different model architectures), and
it achieves the highest pixel accuracy.

\paragraph{Experiment 2: Application in zero-shot manner}
Here, all models are trained on \textit{BlessemFlood21} and tested without further training 
on the \textit{NeuenahrFlood} test set. We first quantitatively inspect zero-shot 
predictions and probability maps of Prithvi-2.0-UPN and then evaluate zero-shot performance 
on the test set at a fixed decision/probability threshold of  $T=0.5$ and under %
variation of $T$.

\begin{figure}[t]
    \centering
    \begin{minipage}[t]{0.24\linewidth}
        \centering
        \includegraphics[width=\linewidth,height=\linewidth,keepaspectratio=false]{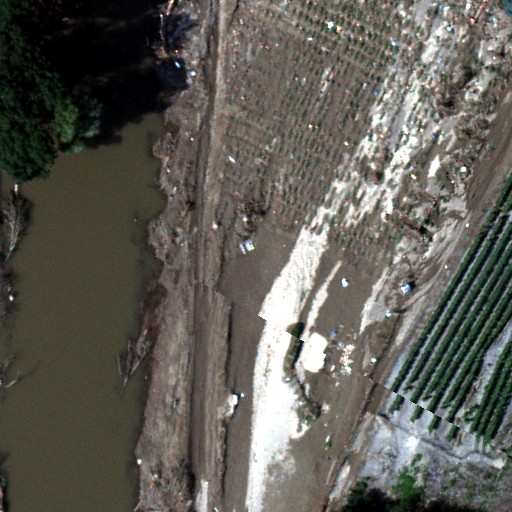}
        \vspace{0.25em}
        \footnotesize (a) RGB
    \end{minipage}
    \hfill
    \begin{minipage}[t]{0.24\linewidth}
        \centering
        \includegraphics[width=\linewidth,height=\linewidth,keepaspectratio=false]{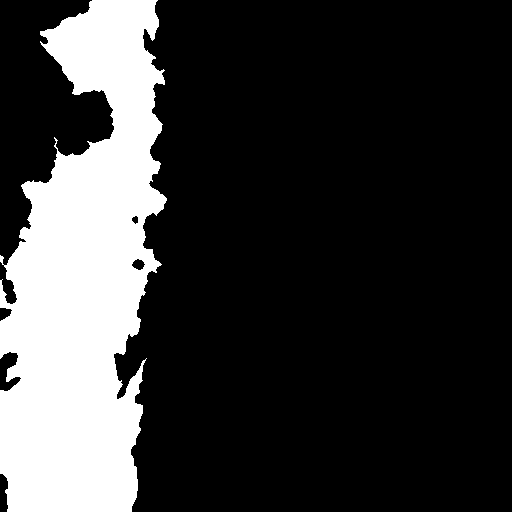}
        \vspace{0.25em}
        \footnotesize (b) Ground truth
    \end{minipage}
    \hfill
    \begin{minipage}[t]{0.24\linewidth}
        \centering
        \includegraphics[width=\linewidth,height=\linewidth,keepaspectratio=false]{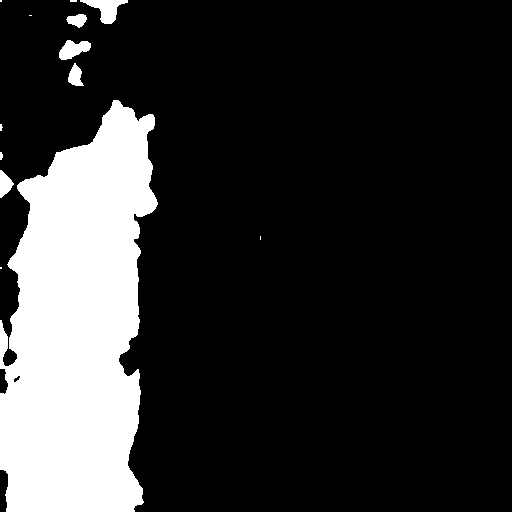}
        \vspace{0.25em}
        \footnotesize (c) $T = 0.02$
    \end{minipage}
    \hfill
    \begin{minipage}[t]{0.24\linewidth}
        \centering
        \includegraphics[width=\linewidth,height=\linewidth,keepaspectratio=false]{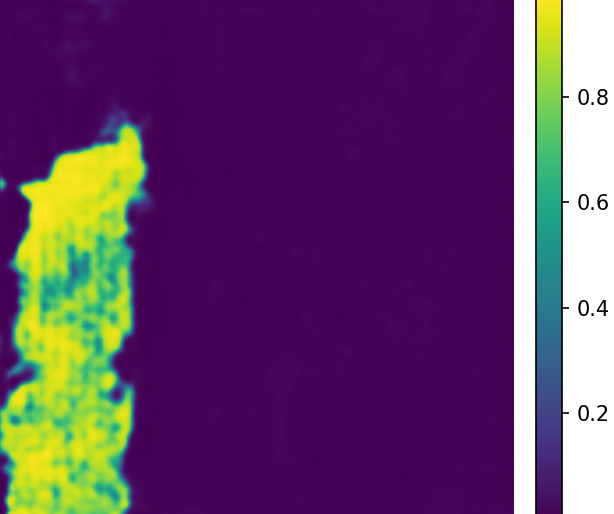}
        \vspace{0.25em}
        \footnotesize (d) Prob. map
    \end{minipage}
\caption{Zero-shot transfer from \textit{BlessemFlood21} to \textit{NeuenahrFlood} (Exp.~2): Prediction of Prithvi-2.0-UPN at the threshold T = 0.02 (best T w.r.t. Dice), and probability map for a NeuenahrFlood test tile.}
\label{fig:neuenahr_zero_vs_finetune_2048}
\end{figure}

Fig.~\ref{fig:neuenahr_zero_vs_finetune_2048} shows a \textit{NeuenahrFlood} test tile with a flooded strip on the right. At a probability threshold of $T = 0.02$, the binary zero-shot prediction in (c) is able to recover large parts of this flooded region. The parts not recovered are hard to detect for the human eye as well. The corresponding probability map in (d) illustrates that rather low thresholds interestingly do not lead to many false positives.

\begin{figure}[ht]
  \centering
  \includegraphics[width=\linewidth,height=0.55\linewidth]{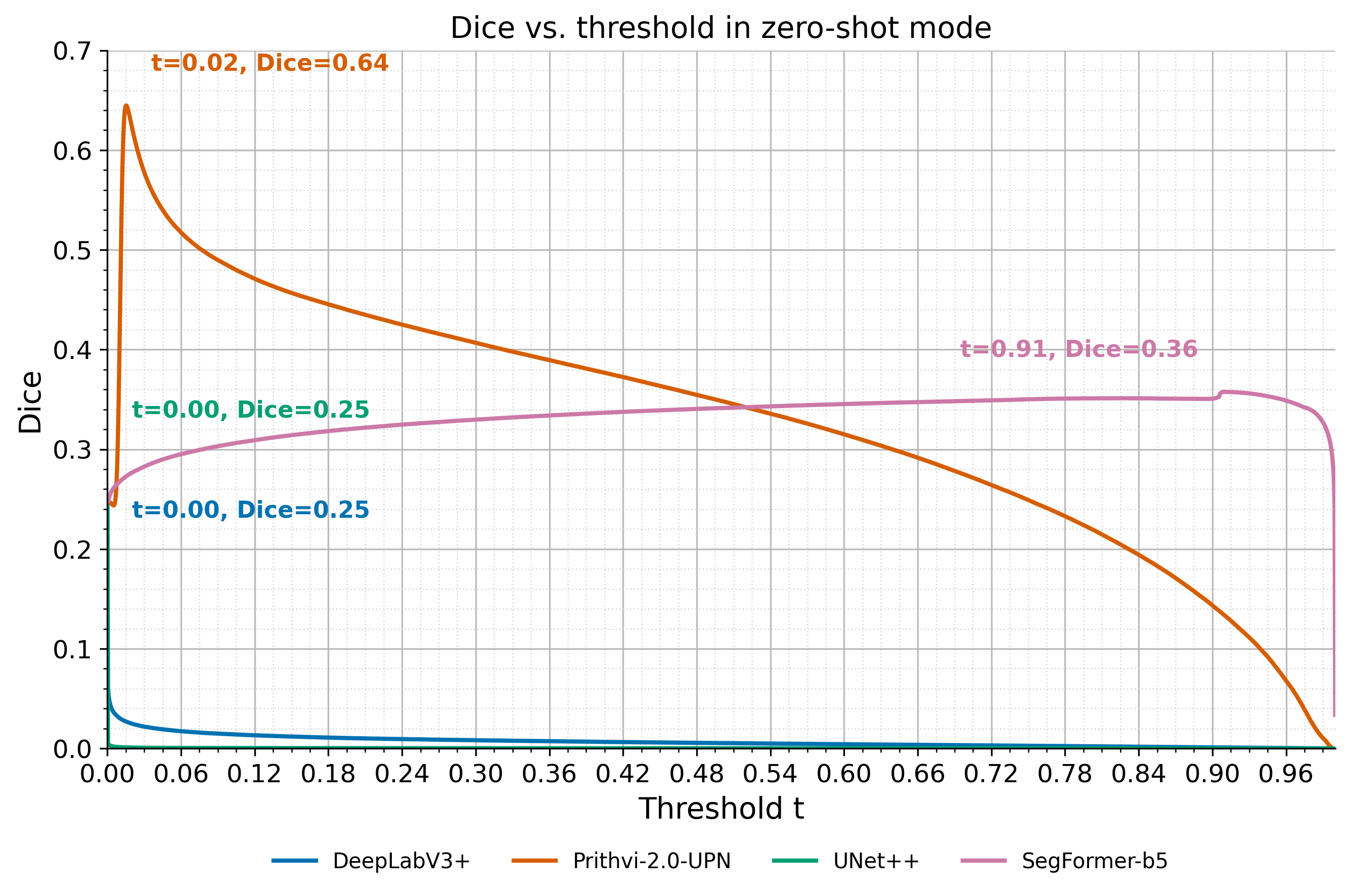}%
\caption{Dice score on \textit{NeuenahrFlood} as a function of the threshold $T$ for models trained on \textit{BlessemFlood21} (Exp.~2).}
  \label{fig:combined_dice_vs_threshold_all_models_v1_default}
\end{figure}

We next quantify zero-shot performance on the \textit{NeuenahrFlood} test data set  at the default threshold $T=0.5$. Tab.~\ref{tab:model_comparison_zeroshot_dice} reports the Dice scores on \textit{BlessemFlood21}. 
DeepLabV3+ and UNet++ reach Dice scores close to zero, whereas SegFormer-b5 and Prithvi-2.0-UPN achieve scores around 34\,\%, with Prithvi-2.0-UPN slightly ahead. 
\begin{table}[ht]
\centering
\small
\setlength{\tabcolsep}{4pt}
\begin{tabular}{l c c c c}
\toprule
                  & DeepLabV3+ & UNet++ & SegFormer-b5 & Prithvi-2.0-UPN \\
\midrule
Dice         & 0.54       & 0.02   & \emph{34.14} & \textbf{34.85}  \\
\bottomrule
\end{tabular}
\caption{Dice score at $T=0.5$ (Exp.~2): models trained on \textit{BlessemFlood21}, tested on \textit{NeuenahrFlood} zero-shot.}
\label{tab:model_comparison_zeroshot_dice}
\end{table}
To assess the dependence on the decision threshold $T$,
we vary it on its whole range from 0 to 1. As metric, we compute the 
Dice score on the \textit{NeuenahrFlood} test set for each model. 
Fig.~\ref{fig:combined_dice_vs_threshold_all_models_v1_default} 
shows the Dice score as a function of the threshold $T$. 
Prithvi-2.0-UPN reaches a maximum Dice score of about 62\,\% at $T = 0.02$, 
and SegFormer-b5 reaches about 36\,\% at $T = 0.91$. 
For DeepLabV3+ and UNet++, Dice is about 25\,\% at $T = 0$, 
which in our setup corresponds to predicting all pixels in the mosaic 
as water, given the fraction of water pixels in the test set. 
As the threshold increases from 0, both curves drop quickly towards 
zero which we attribute to the fact that the probability value for 
the water class is very close to zero for all but a few pixels in the 
test set for these models.

These results show that applying Prithvi-2.0-UPN without any adaptation to \textit{NeuenahrFlood} can still yield Dice scores of about 62\,\% with an appropriate threshold, higher than SegFormer-b5 and the CNN baselines. At the same time, all models remain below the scores achieved with training and testing on the same dataset in Experiment~1, indicating that fine-tuning on NeuenahrFlood still offers potential for improvement, which is examined in Experiment~3.

\paragraph{Experiment 3: Transfer to another flood event by fine-tuning on small data shares.}
Here, we start from the BlessemFlood21-trained Prithvi-2.0-UPN from Experiment~2 and fine-tune the model end-to-end on subsets of the \textit{NeuenahrFlood} training data of increasing size of 128, 256, 512, 1\,024, and 2\,048 tiles. All models are evaluated on the \textit{NeuenahrFlood} test data set (at the natural threshold T = 0.5). The same setup is used for the baseline models.

\begin{figure}[t]
    \centering
    \begin{minipage}[t]{0.24\linewidth}
        \centering
        \includegraphics[width=\linewidth]{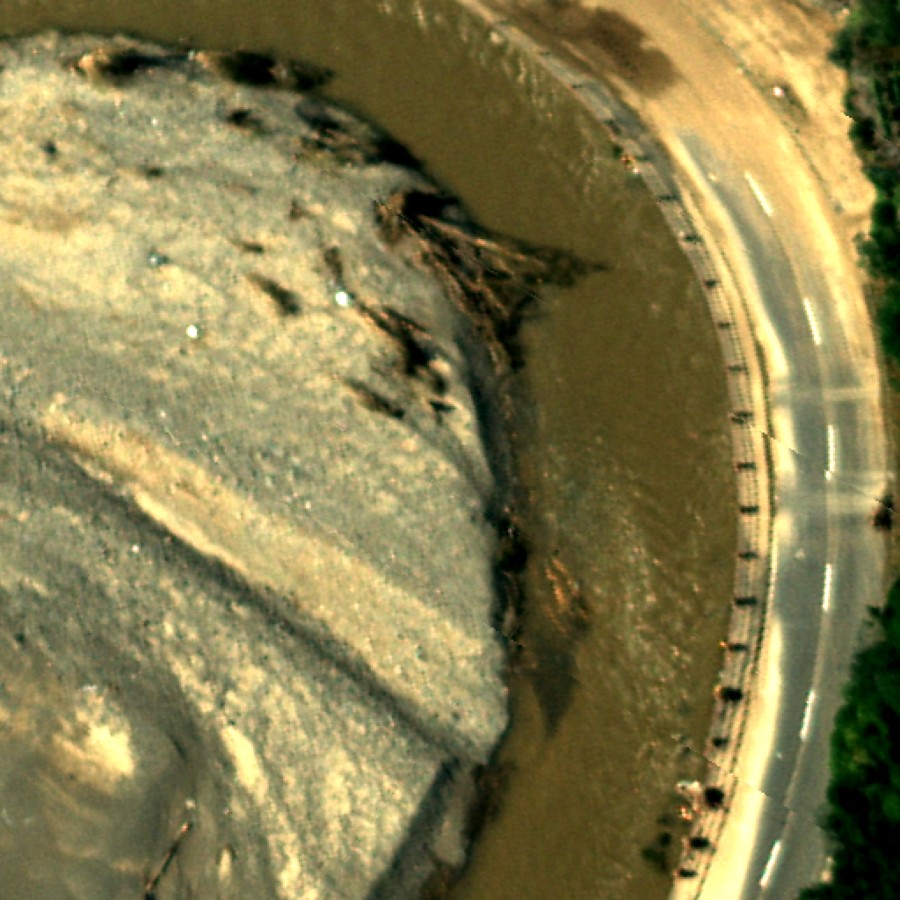}
        \vspace{0.25em}
        \footnotesize (a) RGB input
    \end{minipage}
    \hfill
    \begin{minipage}[t]{0.24\linewidth}
        \centering
        \includegraphics[width=\linewidth]{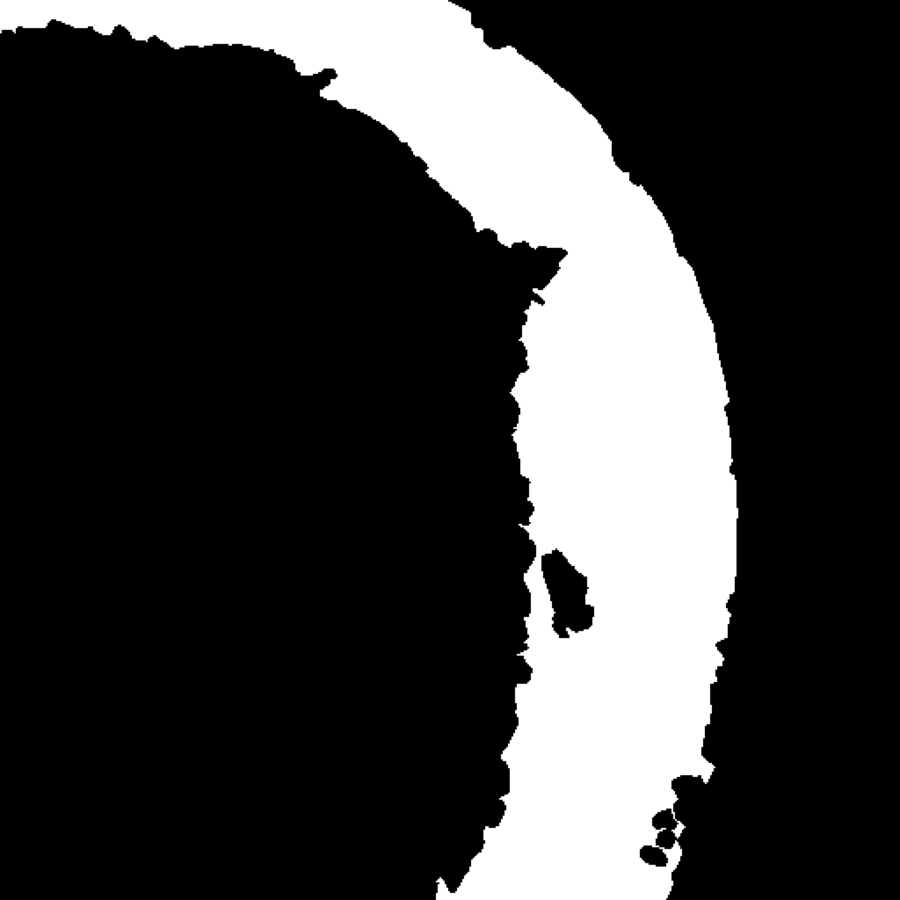}
        \vspace{0.25em}
        \footnotesize (b) Ground truth
    \end{minipage}
    \hfill
    \begin{minipage}[t]{0.24\linewidth}
        \centering
        \includegraphics[width=\linewidth]{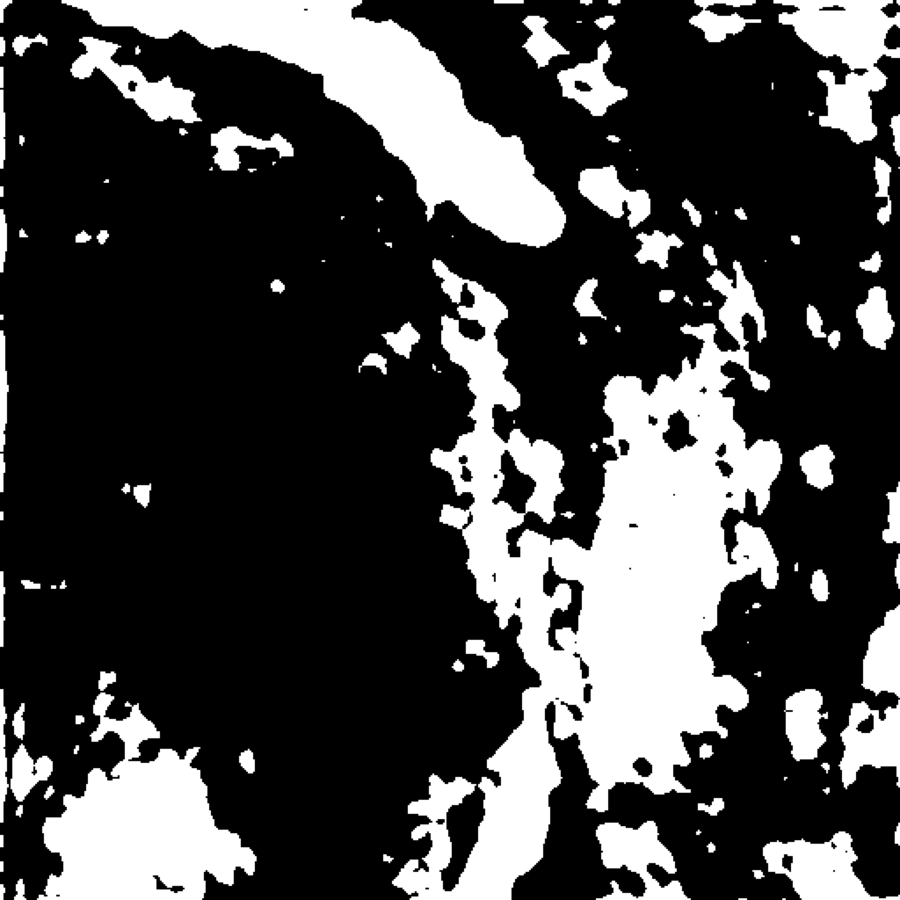}
        \vspace{0.25em}
        \footnotesize (c) Zero-shot
    \end{minipage}
    \hfill
    \begin{minipage}[t]{0.24\linewidth}
        \centering
        \includegraphics[width=\linewidth]{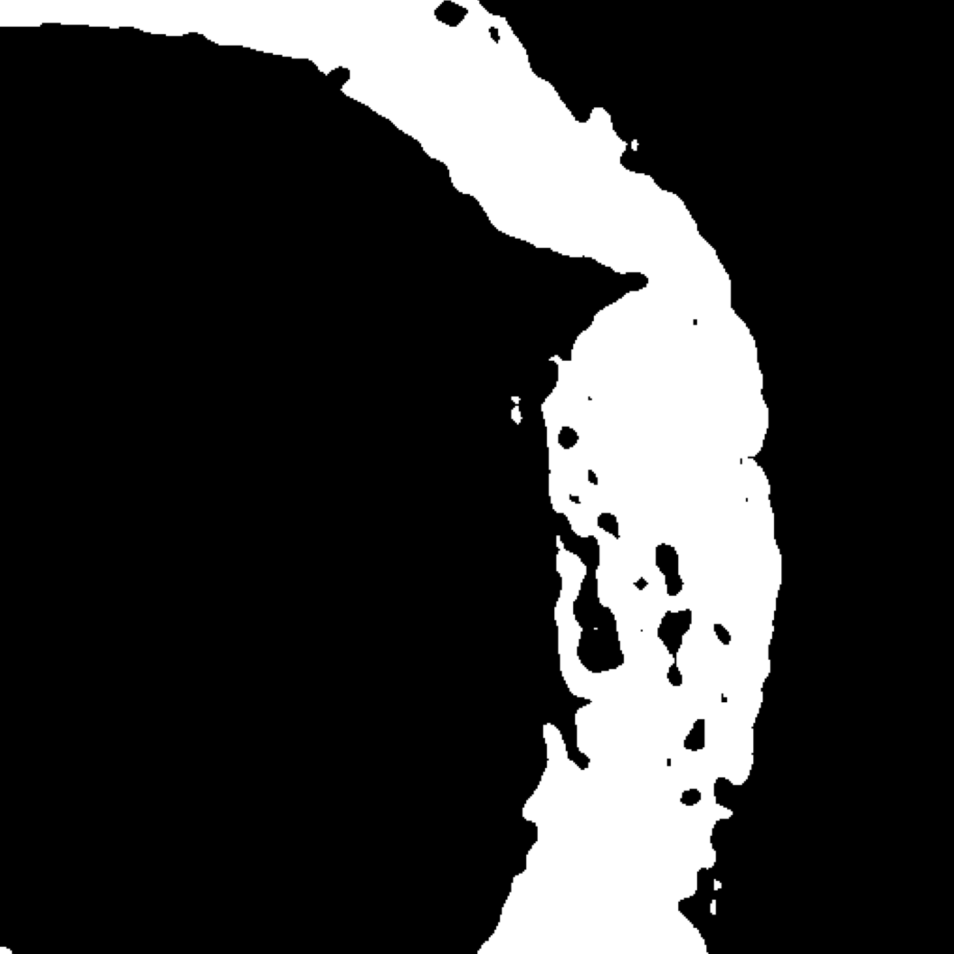}
        \vspace{0.25em}
        \footnotesize (d) Fine-tuned
    \end{minipage}
\caption{Zero-shot prediction vs. prediction after fine-tuning on 2\,048 \textit{NeuenahrFlood} tiles in Exp.~3 (T = 0.5).}
    \label{fig:neuenahr_zero_vs_finetune_2048_tile22}
\end{figure}

Fig.~\ref{fig:neuenahr_zero_vs_finetune_2048_tile22} illustrates how fine-tuning changes the predictions on a curved river reach. The zero-shot prediction at T=0.5 in (c) contains many spurious detections and gaps in the flooded area. After fine-tuning on 2\,048 NeuenahrFlood tiles, the prediction in (d) forms a largely continuous flooded region that follows the annotated boundary more closely and contains fewer isolated pixels. %

\begin{figure}[t]
    \centering
    \includegraphics[width=\columnwidth, height=0.55\linewidth]{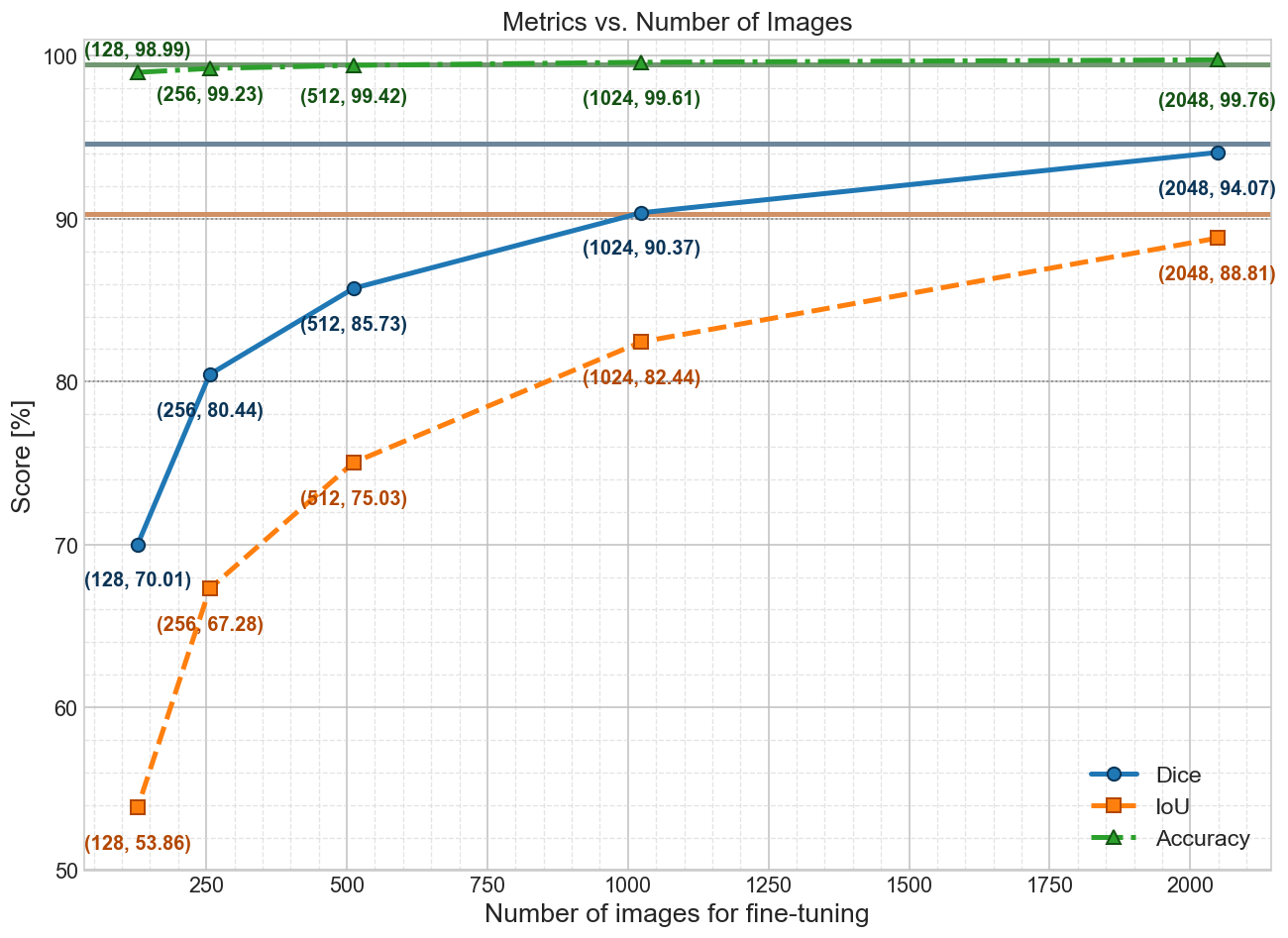}
\caption{Exp.~3: Metrics of Prithvi-2.0-UPN on the \textit{NeuenahrFlood} test set as a function of the number of fine-tuning tiles; horizontal lines correspond to the scores from Exp.~1 (Tab.~\ref{tab:model_comparison_ahr}).}
\label{fig:neuenahr_num_images}
\end{figure}

Fig.~\ref{fig:neuenahr_num_images} summarizes the quantitative evolution of Prithvi-2.0-UPN on 
the~\textit{NeuenahrFlood} test set w.r.t.\ the number of fine-tuning tiles. 
IoU, Dice score, and Accuracy increase with the subset size and move towards the reference levels 
from Exp.~1 (full NeuenahrFlood data set for training). 
With 2\,048 tiles, corresponding to about 9.5\,\% of all NeuenahrFlood training tiles, 
Prithvi-2.0-UPN achieves Dice above 90\,\% at threshold 0.5, together with high IoU and Accuracy.

\begin{table}[t]
\centering
\small
\setlength{\tabcolsep}{4pt}
\begin{tabular*}{\columnwidth}{@{\extracolsep{\fill}} l r r r r r @{}}
\toprule
Model / \# tiles     & 128   & 256   & 512   & 1\,024 & 2\,048 \\
\midrule
Prithvi-2.0-UPN      & 73.52 & 78.44 & 81.12 & 88.01 & 91.34 \\
DeepLabV3+           & 54.37 & 62.44 & 67.89 & 70.20 & 72.31 \\
UNet++               & 44.75 & 55.12 & 61.36 & 66.82 & 69.71 \\
SegFormer-b5         & 66.81 & 70.31 & 74.70 & 78.08 & 80.43 \\
\bottomrule
\end{tabular*}
\caption{Fine-tuning on \textit{NeuenahrFlood} (Exp.~3): Dice on the \textit{NeuenahrFlood} test mosaic vs.\ number of fine-tuning tiles. All models are initialized from \textit{BlessemFlood21}-trained weights.}
\label{tab:dice_finetuning}
\end{table}

Tab.~\ref{tab:dice_finetuning} compares Prithvi-2.0-UPN with DeepLabV3+, UNet++, and SegFormer-b5 
in the same fine-tuning setup. For all subset sizes, Prithvi-2.0-UPN attains the highest Dice scores. 
With 128 tiles it is already ahead of the baselines, and at 2\,048 tiles the gap to the best baseline 
(SegFormer-b5) is about 11 percentage points (91.34\,\% vs.\ 80.43\,\%). 
Prithvi-2.0-UPN gains more Dice score with additional NeuenahrFlood training data than the baseline models.


\section{Conclusion}

We adapted the satellite-pretrained Vision Foundation Model Prithvi-2.0-UPN 
to the task of water segmentation on RGB images
using high-resolution airborne RGB datasets. 
First, we adapted and fine-tuned Prithvi-2.0-UPN 
on the BlessemFlood21 and NeuenahrFlood datasets and assessed its in-domain 
(training and evaluation on the same flood event) performance relative to baseline models.
We have demonstrated that Prithvi-2.0-UPN performs comparably to the baseline models.
Further, we investigated the generalization across flood events in two experiments.
Studying the transfer from BlessemFlood21 to NeuenahrFlood in a zero-shot fashion 
we have demonstrated that Prithvi-2.0-UPN 
improves upon the state-of-the-art, and that there are indications of zero-shot capabilities,
but also that there is room for improvement.
Hence, we have investigated how the performance improves compared with the baseline models
when fine-tuning the models with small shares of NeuenahrFlood training data.
Prithvi-2.0-UPN improved the fastest and reached almost 
the level of performance when fully trained on NeuenahrFlood
indicating transfer capabilities. Future work includes further ablation studies, robustness analysis, as well as the investigation of the impact of automatic parameter adaptation.

\section{Acknowledgments}
The authors gratefully acknowledge the computing time provided 
on the high-performance computer Lichtenberg II at TU Darmstadt, 
funded by the German Federal Ministry of Education and Research (BMBF) 
and the State of Hesse.

\small
\bibliographystyle{IEEEtranN}
\bibliography{references}

\end{document}